\documentclass[sigconf, nonacm]{acmart}

\usepackage{pvldb}

\renewcommand\vldbdoi{XX.XX/XXX.XX}
\renewcommand\vldbpages{XXX-XXX}
\renewcommand\vldbavailabilityurl{}

\usepackage{comment}

\newcommand{\perfresultsline}[5]{%
\begin{figure*}[tb!]
    \centering
    \includegraphics[width=0.85\textwidth]{figures/#1/line/10Percent/Isolet.pdf}
    \Description{Two line plots of #4 and #5 against the percentage of selected features on the Isolet dataset, with one line per feature selection method and a shaded band for the standard deviation of the random baseline.}
    \caption{#4 and #5 of the unsupervised feature selection methods and the random baseline on the Isolet dataset over the first 10\% of features, using #3. The shaded area shows the standard deviation of the random baseline.}
    \label{fg:line-#2}
\end{figure*}
\begin{figure*}[tb!]
    \centering
    \includegraphics[width=0.85\textwidth]{figures/#1/zscore/10Percent/Isolet.pdf}
    \Description{Two line plots of the Z-score of #4 and #5 relative to the random baseline against the percentage of selected features on the Isolet dataset, with one line per feature selection method.}    
    \caption{Z-score of #4 and #5 relative to the random baseline on the Isolet dataset over the extreme dimensionality reduction experiment (0.5\% to 10\%), using #3.}
    \label{fg:z-#2}
\end{figure*}
}

\newcommand{\perfresultsCD}[5]{%
\begin{figure*}[tb!]
    \centering
    \includegraphics[width=0.46\textwidth]{cdd_output/#1/cdd_#4_100Percent_std.pdf}\hfill
    \includegraphics[width=0.46\textwidth]{cdd_output/#1/cdd_#5_100Percent_std.pdf}
    \Description{Two critical difference diagrams showing the average ranks of the feature selection methods for #4 and #5 over the 100\% experiment, with bars connecting methods whose differences are not significant.}
    \caption{Critical difference diagrams of #4 (left) and #5 (right) over the 100\% experiment based on the FSDEM score, using #3.}
    \label{fg:cdd100-#2}
\end{figure*}
\begin{figure*}[tb!]
    \centering
    \includegraphics[width=0.46\textwidth]{cdd_output/#1/cdd_#4_10Percent_std.pdf}\hfill
    \includegraphics[width=0.46\textwidth]{cdd_output/#1/cdd_#5_10Percent_std.pdf}
    \Description{Two critical difference diagrams showing the average ranks of the feature selection methods for #4 and #5 over the first 10\% of features, with bars connecting methods whose differences are not significant.}
    \caption{Critical difference diagrams of #4 (left) and #5 (right) over the first 10\% of features based on the FSDEM score, using #3.}
    \label{fg:cdd10-#2}
\end{figure*}}

\begin{document}
\title{Worse than Random: The Importance of a Baseline for Unsupervised Feature Selection -- Extended Version} 

\author{Muhammad Rajabinasab}
\affiliation{%
  \institution{University of Southern Denmark}
  \city{Odense}
  \country{Denmark}
}
\email{rajabinasab@imada.sdu.dk}

\author{Michael E. Houle}
\affiliation{%
  \institution{New Jersey Institute of Technology}
  \city{Newark}
  \state{NJ}
  \country{United States}
}
\email{michael.houle@njit.edu}

\author{Oussama Chelly}
\affiliation{%
  \institution{Oratio Technologies}
  \city{Tunis}
  \country{Tunisia}
}
\email{oussama.chelly@oratiotechnologies.com}

\author{Arthur Zimek}
\affiliation{%
  \institution{University of Southern Denmark}
  \city{Odense}
  \country{Denmark}
}
\email{zimek@imada.sdu.dk}

\begin{abstract}
Many novel unsupervised feature selection methods are proposed each year, yet their empirical evaluation is limited to supervised and unsupervised evaluation metrics computed on selected datasets, along with comparisons to existing methods. However, in the absence of an established evaluation baseline, it is difficult to determine the value added to the existing literature by each of these methods, and how effective their underlying approaches are. In this evaluation study, we are using random feature selection as a baseline for evaluating unsupervised feature selection methods. While a random selection of features might seem to be a rather weak baseline, we show in this evaluation study that many of the state-of-the-art methods in unsupervised feature selection are outperformed by random feature selection in both performance and efficiency. Accordingly, we emphasize the strict requirement of considering random feature selection as a baseline in the development process of novel unsupervised feature selection methods, and we encourage the community to develop better evaluation measures for feature selection. Both would help to ensure consistent improvement and realistic assessment.
\end{abstract}

\maketitle

\vldbtopmatter 

\section{Introduction}
Evaluation baselines are essential for machine learning applications to provide a point of comparison against which the performance of a complex model can be measured. Typically, evaluation baselines are simple non-ML models, or very basic ML models which help to assess how a realistic na\"{i}ve model would perform on a specific task and dataset. For instance, in binary classification, a constant function which always predicts the majority class can be considered as a baseline to evaluate classification models. In regression, a simple linear regression model often acts as a baseline.

With ongoing research over time, it is unavoidable that a different set of methods is evaluated against each other on a different collection of datasets in different studies. But it is important to notice that performance quality comparisons are not transitive if the datasets (and potentially hyperparameters) are not the same. Method $B$ can perform better than Method $A$ on dataset collection $X$ and testbed $\theta_X$, and method $C$ can perform better than method $B$ on dataset collection $Y$ and testbed $\theta_Y$, yet we cannot conclude anything about the comparison between methods $C$ and $A$ on $X$, $Y$, or other datasets, or other settings. Therefore, without using a proper baseline, it is impossible to justify if a complex opaque model is truly adding value. If a complex model is better than a trivial baseline only by a small margin, its complexity and computational cost might be considered unjustified. A proper baseline helps identify the difficulty of the task and ensures that the achieved gains in the performance are actually meaningful and not just the result of a random coincidence or of a simplistic dataset structure.

Feature selection is the process of selecting the most important features which represent the data the best, and of removing the redundancies. As a data reduction step, feature selection lowers the storage, memory, and computational cost of all downstream analytics in a data pipeline, and often improves their quality. Feature selection is often employed in a supervised setting, where either statistical values are calculated as feature importance (filter methods) or feature importance is assessed based on the impact of the feature on a classifier’s prediction (wrapper methods) \citep{DBLP:journals/apin/DhalA22}. Recently, unsupervised feature selection methods have gained attention for their ability to select the most informative features without requiring the existence of labels, as those are often not available at all or at a stage when the feature set is to be reduced \citep{10123301, DBLP:journals/ijon/ShangKWZWLJ23, DBLP:journals/prl/WANG2024110183}.

For supervised feature selection, simple in\-for\-ma\-tion-the\-o\-ret\-ic methods such as Mutual Information (MI) \citep{peng2005feature, cai2020feature} which measure the information gain regarding the class label obtained by observing the values of a specific feature, can serve as a suitable baseline. We can denote methods such as Variance-based \citep{guyon2003introduction} and Correlation-based \citep{hall1999correlation, mitra2002unsupervised} as similarly simple methods for unsupervised feature selection. However, these methods may not be suitable as baselines for feature selection in many data types, such as images, where the variance of each feature (pixel) appears  equal, or data with signals from independent sources. Moreover, in very high-dimensional spaces, arguably one of the most important domains to apply feature selection algorithms on, their performance drops significantly, transforming them into an extremely na\"ive baseline, similar to using random chance (50\%) as a baseline for binary classification when the majority of the samples belong to one of the classes.  

In this paper, we use random feature selection as a baseline for the evaluation of unsupervised feature selection methods. Random feature selection is the process of randomly sorting the features to provide feature ranking and importance for conducting the feature selection task. Clearly, random feature selection does not require any labels and hence can be deemed as unsupervised. It also does not require any expensive computational steps and therefore is very efficient. On the other hand, in very high-dimensional spaces where a small proportion of features can still discriminate data points from each other, it is expected to offer an acceptable overall performance, advertising itself as a proper candidate to be the baseline for the evaluation of unsupervised feature selection methods. 

We conduct an extensive experimental study of classic and state-of-the-art unsupervised feature selection methods against the random baseline on 23 high-dimensional benchmark datasets, covering both efficiency and downstream performance with five classifiers and $k$-means clustering, two ranges of selected features, and a statistical analysis over all datasets~\footnote{Python implementation is available on: \url{https://github.com/mrajabinasab/worse_than_random/}}. Our main findings are as follows:
\begin{itemize}
    \item None of the evaluated methods is significantly better than the random baseline, for any of the five classifiers or for clustering, and the random baseline is ranked among the top five methods in every comparison.
    \item Several methods, including the recent state-of-the-art methods VCSDFS and SOGFS, are ranked behind the random baseline in almost all comparisons.
    \item The methods ranked ahead of the random baseline depend on the downstream task: SCFS and MCFS are the best methods for classification, and LLSRFS and Laplacian Score are the best methods for clustering. However, none of them is significantly better than the random baseline in any setting. Additionally, methods such as SCFS or VCSDFS are advertised as unsupervised, while they take advantage of the number of classes (i.e., supervised information) in the selection process.
    \item The random baseline is (unsurprisingly) the most efficient by a large margin, while the most expensive state-of-the-art methods show high runtime even on moderately sized data.
    \item The performance of the random baseline varies little across independent runs, and the competing methods more often lie clearly below this variation than clearly above it. However, a single random selection is theoretically not reliable when only a few features are selected, and the mean and the standard deviation over many runs are required.
    \item The traditional quality measures for feature selection are not able to detect random-level quality. They can be used in a relative way only, which requires a meaningful baseline (such as random feature selection) to give meaningful insights on the quality level of competitors.
\end{itemize}

The rest of the paper is structured as follows: We provide a brief literature review on unsupervised feature selection in Section~\ref{sec:lr}, focusing on recent methods and their evaluation strategies. We discuss the methodology behind using random feature selection as a baseline for the evaluation of unsupervised feature selection methods in Section~\ref{sec:meth}. In Section~\ref{sec:eval}, we provide the details of the experimental setup used for the empirical evaluation conducted in the paper. We present the experimental results in Section~\ref{sec:evr}, providing an in-depth experiment on established and state-of-the-art unsupervised feature selection methods in comparison with the random baseline, including their efficiency, their supervised and unsupervised performance, and the variability of the random baseline, discussing the results and highlighting the shortcomings of recent work in the literature. Finally, in Section~\ref{sec:conc}, we present of our findings and their consequences. 

\section{Literature Review}\label{sec:lr}
Unsupervised feature selection is an important topic in data mining and machine learning which has been studied over many decades. On the more established and traditional side, methods such as Variance-based \citep{guyon2003introduction} and Correlation-based \citep{hall1999correlation, mitra2002unsupervised} are considered the simplest ways of conducting unsupervised feature selection. The unsupervised general-purpose forward-filter 
Laplacian Score (LS) feature selection method~\citep{laplacianscore}
is one of the most popular. It belongs to the general framework of spectral feature selection~\citep{spectral},
and has a computational complexity of $O(dn^{2})$, where $d$ is the dimensionality (number of features) and $n$ is the number of samples. Unsupervised feature selection can also be based on
how well the selected features preserve the cluster structure of the data. This is the approach of Multi-Cluster Feature Selection (MCFS)~\citep{mcfs}.

Despite the fact that feature selection is one of the classical problems in the field of machine learning and data mining, it is still the subject of many research papers in recent years. 

Subspace learning, cluster analysis and sparse learning are utilized for unsupervised feature selection (SCFS) \citep{DBLP:journals/eaai/ParsaZG20}, and a self-expressive model is employed to learn cluster similarities. A~regularized regression approach is used to capture the existing correlations among features and clusters sparsely. 
SOGFS \citep{DBLP:conf/aaai/WuC21} performs feature selection and local structure learning simultaneously.  
LLSRFS \citep{DBLP:journals/prl/WANG2024110183} introduces an exponential weighting mechanism to adjust feature weight distribution. 
Neural Networks Embedded Self-expression (NNSE) \citep{DBLP:journals/pr/YouYHL23} utilizes neural networks and embeds them into the self-expression model in order to enhance the representative ability by preserving the local structure with an adaptive graph regularization module. Variance–covariance subspace distance (VCSDFS) \citep{DBLP:journals/nn/KaramiSTMV23} utilizes the correlation of information included in the features of data, thus determining all the feature subsets whose corresponding Variance–Covariance matrix has the minimum norm property.  
Robust, Adaptive and Flexible Graph (RAFG) \citep{Jiang2024} is a graph-learning framework proposed for unsupervised feature selection. The $L_{2,1}$-norm is imposed on the flexible regression term to alleviate the adverse effects of both noisy features and outliers, and an $L_{2,1}$-norm regularization term is incorporated to ensure that the selected transformation matrix is sufficiently sparse. Most of the proposed methods in recent years, however, have focused on the unsupervised multi-view feature selection problem \citep{yang2025tensor,  cao2024structure, wu2024collaborative}.

The evaluation of unsupervised feature selection methods is often based on a limited selection of datasets. On the other hand, the whole concept of evaluation is based on a comparison with other existing methods, and the lack of an evaluation baseline is the common shortcoming of these papers. Some of the papers, such as the work of \citet{mcfs}, use the performance with all of the features as a baseline. However, in very high-dimensional spaces, there is arguably a high amount of redundancy among features and many of the features act as noise. Therefore, it is not difficult to beat such a baseline. Consequently, a new method can outperform its competitors and still be no better than a random selection of features, which is not revealed by a comparison with existing methods alone.

Only a few papers focus on the evaluation of feature selection algorithms. \citet{DBLP:conf/pkdd/Nogueira016} propose a method to measure the stability of the feature selection algorithms in the presence of noise. Baseline Fitness Index (BFI) \citep{mostert2021feature} combines the amount of feature selection and the performance as a single measure. \citet{Rajabinasab2024fsdem} propose a way to evaluate the overall quality of the feature selection process and the stability of the feature selection algorithms based on the gain of adding more features.

Several studies benchmark feature selection methods empirically. \citet{bommert2020benchmark} compare 22 filter methods on 16 high-dimensional datasets and use the performance without a baseline. \citet{solorio2020review} review unsupervised feature selection methods and their evaluation practices. \citet{kim2026evaluation} show that the ranking of unsupervised feature selection methods changes with the choice of the evaluation setting. None of these works, however, offers a baseline to guide the development and the evaluation of unsupervised feature selection methods. 

Random baselines have proven to be strong competitors in other areas of machine learning. Random search was shown to find hyperparameters as good as or better than grid search at a fraction of the computational cost \citep{bergstra2012random}, and random search was found to be a competitive baseline for neural architecture search, often on par with sophisticated search strategies \citep{li2019random, yang2020nas}. Moreover, random subsets of features are used to construct ensembles of classifiers, as in the random subspace method \citep{ho1998random}. These findings motivate a closer look at a random baseline for unsupervised feature selection.

\section{Methodology}\label{sec:meth}
In this study, we evaluate some classic as well as recent state-of-the-art unsupervised feature selection methods with random feature selection as a baseline, with traditionally used evaluation measures. We run random feature selection 100 times and take the average value of the evaluation metrics as the reference value for the random baseline. Highlighting the mean and the standard deviation of these values in the plots depicting the evaluations results also helps with assessing how well an unsupervised feature selection algorithm is performing in comparison to the random baseline. 

Given a dataset $\mathcal{D} = \{\mathbf{x}_i\}_{i=1}^n$ with $n$ instances and $d$ features, the feature selection objective is to select a subset of $k$ features, $\mathcal{F}_k \subset \mathcal{F}_d$, where $\mathcal{F}_d$ is the set of all $d$ features and $k < d$. The random feature selection operates by assigning a feature importance score $s_j$ to each feature $j \in \{1, \dots, d\}$, drawn independently from the uniform distribution $\mathcal{U}(0,1)$. The vector of scores is $\mathbf{s} = [s_1, s_2, \dots, s_d]$, and the $k$ features are selected by taking the top $k$ features corresponding to the largest scores in $\mathbf{s}$.
The set of selected feature indices $\mathcal{I}_k$ is defined as:
\begin{equation}
\mathcal{I}_k = \operatorname{Top}_k(\mathbf{s}) = \{\, j \mid s_j \geq s_{(k)} \,\},
\end{equation}
where $s_{(k)}$ denotes the $k$-th largest score. As the scores are independent and identically distributed, every subset of $k$ features is selected with the same probability $1/\binom{d}{k}$. Moreover, as $\mathbf{s}$ defines a complete ranking of the features, the selected subsets are nested for increasing $k$, as for any ranking-based feature selection method.

Let us note that, technically, a simple shuffling and selection of the $k$ first features is all that is needed for random feature selection. However, we present the method based on feature scores to match the framework of regular feature selection algorithms.

The random baseline is the expected performance of a randomly selected subset of features. For a performance measure $P$ and $k$ selected features, it is estimated over $R$ independent runs as
\begin{equation}\label{eq:mean}
\mu_{random}(k) = \frac{1}{R}\sum_{r=1}^{R} P\big(\mathcal{I}_k^{(r)}\big),
\end{equation}
\begin{equation}\label{eq:stddev}
\sigma_{random}(k) = \sqrt{\frac{1}{R-1}\sum_{r=1}^{R}\Big(P\big(\mathcal{I}_k^{(r)}\big) - \mu_{random}(k)\Big)^2},
\end{equation}
where $\mathcal{I}_k^{(r)}$ is the subset selected in run $r$. We use $R=100$. The mean serves as the reference value of the baseline, and the standard deviation quantifies the variation over a set of random selections.

We anticipate the presence of many redundant features in very high-dimensional spaces. Hence, by even randomly removing features, the evaluated downstream application is still expected to be successful. Conversely, if only a few features carry the relevant information, a random subset is unlikely to contain them, and this is exactly the setting in which a feature selection method should clearly outperform the random baseline. Random feature selection is also clearly efficient as it only requires generating some random values as feature importance scores (or just shuffling). Its cost is at most $O(d \log d)$ for ranking the $d$ scores, independent of the number of instances $n$. As the most na\"ive and still logical solution to the unsupervised feature selection problem, random feature selection can be considered as a suitable baseline for the evaluation. Our results will demonstrate that such a baseline should indeed be considered indispensable for the development and evaluation of unsupervised feature selection algorithms.

\section{Experimental Setup}\label{sec:eval}
In this section, we present the experimental setup to evaluate the feature selection performance of various unsupervised feature selection methods in comparison with the random baseline. The experiments include traditional methods such as  Variance-based \citep{guyon2003introduction}, Correlation-based \citep{hall1999correlation, mitra2002unsupervised}, Laplacian Score~\citep{laplacianscore}, and MCFS~\citep{mcfs}, as well as recent state-of-the-art methods including SCFS \citep{DBLP:journals/eaai/ParsaZG20}, SOGFS \citep{DBLP:conf/aaai/WuC21}, LLSRFS \citep{DBLP:journals/prl/WANG2024110183}, and VCSDFS \citep{DBLP:journals/nn/KaramiSTMV23}. Our goal is not an exhaustive comparison of all existing methods, but to show that established as well as recent methods do not necessarily outperform the random baseline, and that the random baseline is therefore a necessary point of comparison in the development of new methods.

\subsection{Benchmark Datasets}
To evaluate our approach, we perform extensive experiments on 23 high-dimensional datasets from the scikit-feature repository \citep{fs_datasets}. These datasets are widely recognized benchmarks in feature selection research \citep{DBLP:journals/eaai/ParsaZG20, DBLP:journals/nn/KaramiSTMV23, mcfs}. Table~\ref{tbl:hdds} lists their characteristics. They cover gene expression, image, speech, mass spectrometry, and synthetic data, and in 19 of the 23 datasets the number of features exceeds the number of instances.

\begin{table}[tb]
    \centering
    \caption{Characteristics of the benchmark datasets.}
    \label{tbl:hdds}
    \small
    \setlength{\tabcolsep}{3pt}
    \begin{tabular}{|l|r|r|r|l|}
        \hline
        \textbf{Dataset} & \textbf{Instances} & \textbf{Features} & \textbf{Classes} & \textbf{Domain} \\
        \hline
        ALLAML & 72 & 7,129 & 2 & Biological \\
        arcene & 200 & 10,000 & 2 & Mass spec. \\
        Carcinom & 174 & 9,182 & 11 & Biological \\
        CLL\_SUB\_111 & 111 & 11,340 & 3 & Biological \\
        COIL20 & 1,440 & 1,024 & 20 & Image \\
        colon & 62 & 2,000 & 2 & Biological \\
        gisette & 7,000 & 5,000 & 2 & Digits \\
        GLI\_85 & 85 & 22,283 & 2 & Biological \\
        GLIOMA & 50 & 4,434 & 4 & Biological \\
        Isolet & 1,560 & 617 & 26 & Speech \\
        leukemia & 72 & 7,070 & 2 & Biological \\
        lung & 203 & 3,312 & 5 & Biological \\
        lung\_discrete & 73 & 325 & 7 & Biological \\
        madelon & 2,600 & 500 & 2 & Artificial \\
        ORL & 400 & 1,024 & 40 & Image \\
        orlraws10P & 100 & 10,304 & 10 & Image \\
        pixraw10P & 100 & 10,000 & 10 & Image \\
        Prostate\_GE & 102 & 5,966 & 2 & Biological \\
        SMK\_CAN\_187 & 187 & 19,993 & 2 & Biological \\
        TOX\_171 & 171 & 5,748 & 4 & Biological \\
        warpAR10P & 130 & 2,400 & 10 & Image \\
        warpPIE10P & 210 & 2,420 & 10 & Image \\
        Yale & 165 & 1,024 & 15 & Image \\
        \hline
    \end{tabular}
\end{table}

\subsection{Evaluation Metrics and Methods}
To thoroughly evaluate the feature selection methods across both supervised and unsupervised downstream tasks, four key metrics are employed: Accuracy (ACC), Area Under the Curve (AUC), Clustering Accuracy (CLSACC), and Normalized Mutual Information (NMI). We also use FSDEM and Z-score for further analysis and insights.

\paragraph{Accuracy (ACC)}
Accuracy is a fundamental metric for supervised classification tasks, representing the proportion of the total number of predictions that were correct. For a set of $N$ instances with true labels $y_i$ and predicted labels $\hat{y}_i$, it is calculated as:
\begin{equation}
\text{ACC} = \frac{1}{N} \sum_{i=1}^{N} \mathbb{I}(y_i = \hat{y}_i)
\label{eq:acc}
\end{equation}
where $\mathbb{I}(\cdot)$ is the indicator function.

\paragraph{Area Under the Curve (AUC)}
AUC measures the ability of a classifier to distinguish between classes. It is the area under the Receiver Operating Characteristic (ROC) curve, which plots the True Positive Rate (Sensitivity) against the False Positive Rate (1 - Specificity) at various threshold settings. AUC ranges from 0 to 1, with 0.5 indicating random performance of the classifier and a higher value indicating better performance. For multi-class problems, AUC is computed in a one-vs-rest manner and averaged over the classes.

\paragraph{Clustering Accuracy (CLSACC)}
Clustering Accuracy assesses how well the clusters found by the algorithm match the true labels. It requires finding the best one-to-one mapping (permutation $\pi$) between the cluster assignments ($c_i$) and the true labels ($l_i$).
\begin{equation}
\text{CLSACC} = \frac{1}{N} \sum_{i=1}^{N} \mathbb{I}(l_i = \pi(c_i))
\label{eq:clsacc}
\end{equation}
where $N$ is the number of data points. The optimal mapping $\pi$ is typically found using the Hungarian algorithm \citep{kuhn1955hungarian}.

\paragraph{Normalized Mutual Information (NMI)}
NMI is another metric based on unsupervised tasks that quantifies the degree of dependence between the clustering ($C$) and the true labels ($L$) by normalizing the Mutual Information ($\text{MI}$), making it comparable across different cluster numbers. NMI is calculated as:
\begin{equation}
\text{NMI}(C, L) = \frac{2\,\text{MI}(C, L)}{H(C) + H(L)}
\label{eq:nmi}
\end{equation}
where $\text{MI}(C, L)$ is the Mutual Information between $C$ and $L$, and $H(C)$ and $H(L)$ are the entropies of $C$ and $L$, respectively. 

\paragraph{Feature Selection Dynamic Evaluation Metric (FSDEM)}
To evaluate the overall process of feature selection, we use FSDEM \citep{Rajabinasab2024fsdem} combined with the discussed metrics. FSDEM is calculated as: 
\begin{equation}
    \label{eq:fsdem}
    \text{FSDEM} = \frac{\int_{a}^{b} g(x) \, dx}{b - a}
\end{equation}
where $g(x)$ indicates the approximated function based on different observations of an arbitrary evaluation metric, and $a$ and $b$ indicate the specific range of values in which the performance is evaluated. FSDEM allows us to assess the quality of the overall feature selection process, instead of the individual points. As the observations are equally spaced, we approximate the integral by the average over all observations.

\paragraph{Z-score}
To further analyze the magnitude of the difference between the benchmark methods and the random baseline, we calculate the $Z$-score for each method relative to the distribution of the random baseline's results. The Z-score is defined as:
\begin{equation}
Z = \frac{P_{\mathit{method}} - \mu_{\mathit{random}}}{\sigma_{\mathit{random}}}
\label{eq:zscore}
\end{equation}
where $P_{\mathit{method}}$ is the performance metric of a specific feature selection algorithm, and $\mu_{\mathit{random}}$ and $\sigma_{\mathit{random}}$ are the mean (Eq.~\ref{eq:mean}) and standard deviation (Eq.~\ref{eq:stddev}) of the random baseline's performance, respectively. This transformation allows us to observe how many standard deviations above or below the expected random performance a method performs.

\subsection{Evaluation Configuration}
We perform our experiments using the \texttt{FSEVAL} benchmarking suite \citep{rajabinasab2026fsevalfeatureselectionevaluation}, a specialized framework designed for the extensive and comprehensive evaluation of feature selection algorithms. The supervised evaluation is done using five classifiers, namely KNN \citep{cover1967nearest}, decision tree \citep{breiman1984classification}, SVM \citep{cortes1995support}, random forest \citep{breiman2001random}, and XGBoost \citep{chen2016xgboost}, with a 5-fold stratified cross-validation to provide a robust evaluation. As the results are very similar for all classifiers, we present the results of the random forest classifier in detail and summarize the remaining classifiers in the text.
All remaining results are provided in Appendix~\ref{app:ext}.

For the unsupervised evaluation, we use the average of 10 $k$-means runs, with the number of clusters = number of classes. For the methods that require the number of clusters, i.e., SCFS, VCSDFS, and MCFS, it is likewise set to the number of classes, and Laplacian Score uses a $k$-nearest neighbor graph with $k=5$. All parameters are set to the default values recommended by their authors.

We conduct two different experiments to evaluate the performance of the feature selection process. Firstly, we study the overall feature selection process by selecting from 5\% to 100\% of features with a step size of 5\%. As the datasets are very high-dimensional, we also conduct a different experiment by selecting from 0.5\% to 10\% of features with a step size of 0.5\%. This allows us to have a better and more realistic view of the performance of feature selection algorithms to make sure that a high number of selected features relative to the number of instances does not obscure the results. In realistic feature selection problems, we usually aim to select a very small number of features for the final downstream task. The second experiment allows us to assess the performance of a feature selection method for extreme dimensionality reduction cases.

 We run the experiments on Ubuntu with an AMD Zen 5 CPU and 64GB of RAM. This hardware, however, is not the minimum requirement for running these experiments, and in principle, the experiments can be executed on any modern computer system.

\section{Experimental Results}\label{sec:evr}
We conduct experiments on the selected unsupervised feature selection methods from both performance and efficiency perspectives to highlight the gain and the cost of using complex methods for unsupervised feature selection.

\subsection{Efficiency Analysis}
\begin{figure}[tb!]
    \centering
    \includegraphics[width=\columnwidth]{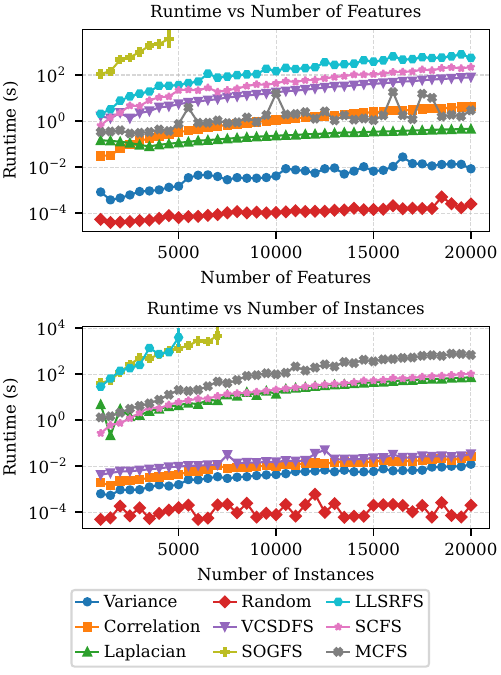} 
    \caption{Runtime analysis of the feature selection methods. The $y$-axis is presented in logarithmic scale.}
    \label{fg:runtime}
\end{figure}

The first experiment presents the runtime of the unsupervised feature selection algorithms (Fig.~\ref{fg:runtime}). For this experiment, we consider two variables, the number of instances and the number of features. We present two results on synthetic data drawn from a standard normal distribution, in which one of the variables is fixed at 100 and the other variable varies from 1000 to 20000 with a step size of 500. For each method, the runtime is capped at one hour. The first time it takes more than one hour for a method to finish the feature selection process, the time is recorded and further investigation is skipped. The $y$-axis is presented in logarithmic scale.

As expected, the random baseline is the most efficient method by a large margin as it does not involve any significant computational operations. It is followed by the traditional methods such as Variance-based and Correlation-based feature selection. Recent state-of-the-art methods are computationally more expensive.

\begin{figure*}[tb]
    \centering
    \includegraphics[width=0.9\textwidth]{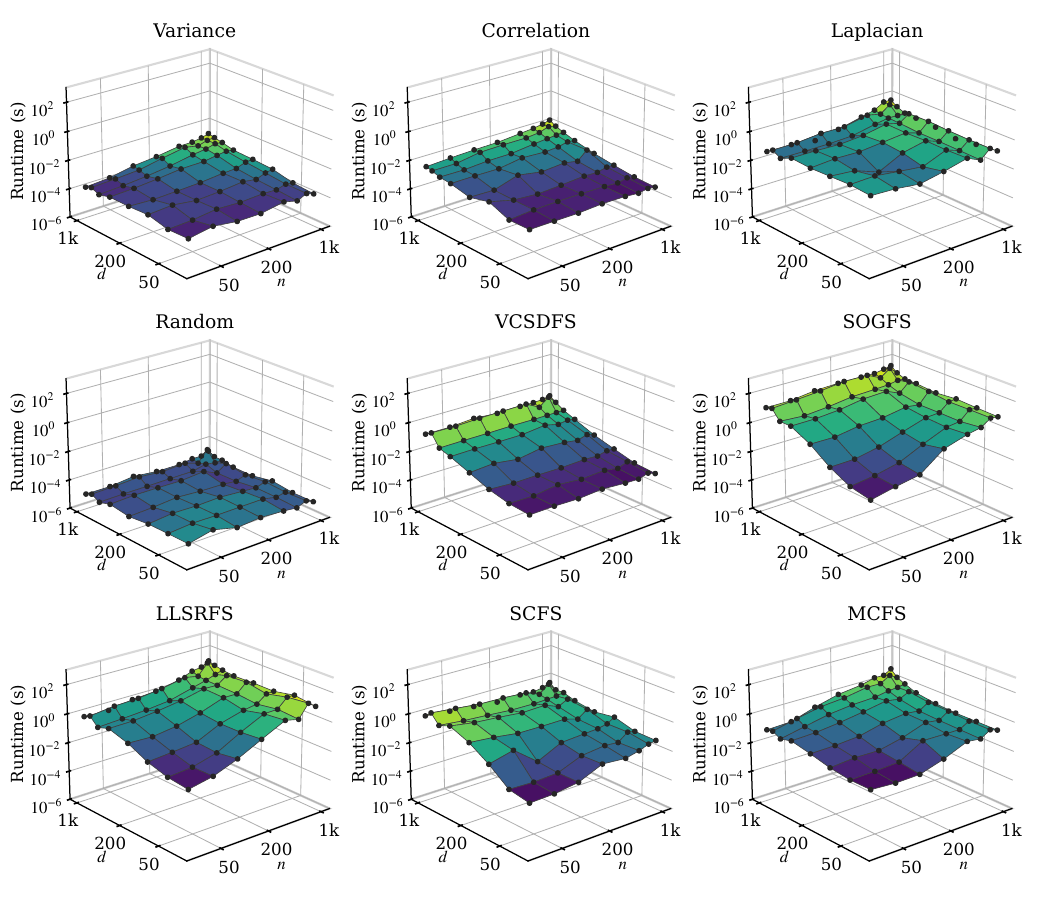}
    \Description{Nine 3D surface plots, one per feature selection method, showing the runtime on a logarithmic scale over the number of instances and the number of features, both between 25 and 1,000.}
    \caption{Runtime of the unsupervised feature selection methods on random data for all combinations of the number of instances $n$ and the number of features $d$ between 25 and 1,000. The runtime axis is logarithmic and shared by all panels, and the colors are scaled per method.}
    \label{fg:runtime-grid}
\end{figure*}

SOGFS and LLSRFS are by far the most expensive methods. In the efficiency analysis, where the other dimension is fixed to 100, SOGFS exceeds the time limit of one hour already at 4,500 features, and LLSRFS at 5,000 instances, whereas 12 of the 23 datasets contain more than 4,500 features and gisette contains 7,000 instances. To verify that this is not an artifact of varying only one dimension at a time, Fig.~\ref{fg:runtime-grid} shows the runtime of the methods on random data for all combinations of $n$ and $d$ between 25 and 1,000, so that the largest configuration contains $10^6$ entries. Even at this scale, SOGFS and LLSRFS are the most expensive methods, and their runtime grows fastest with both $n$ and $d$. This scale is small compared to the benchmark datasets, as 22 of the 23 datasets exceed it in at least one dimension, and the three largest datasets, gisette, SMK\_CAN\_187, and arcene, contain about 35, 3.7, and 2 times as many entries as the largest grid configuration. Hence, on the benchmark datasets, the cost of these two methods is far beyond that of all other methods --- unjustified cost as these methods are not showing better quality.

\perfresultsline{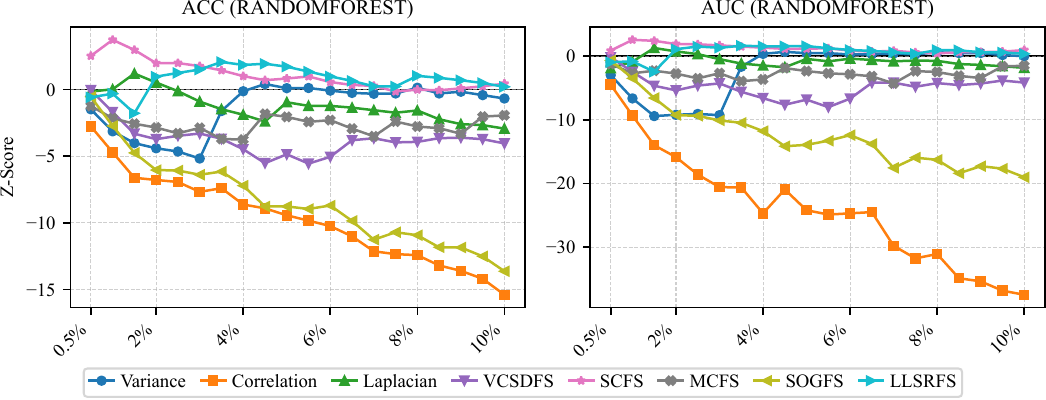}{rf}{the random forest classifier}{ACC}{AUC}

\subsection{Supervised Performance}
We conduct an extensive performance analysis based on our evaluation metrics and on all 23 benchmark datasets. For the supervised metrics (ACC and AUC), we present the results for the random forest classifier in detail and summarize the differences for the remaining classifiers. We show the evaluation results on the Isolet dataset for the 10\% range experiment, the corresponding Z-scores relative to the random baseline, and the critical difference diagrams over all datasets. The results for all datasets and all tested classifiers are provided in Appendix~\ref{app:ext}.

We selected the Isolet dataset for demonstration, as it shows a pattern commonly observed in many other datasets, in which most of the feature selection methods fail to outperform the random baseline. However, the feature selection performance varies on different datasets. Therefore, we also analyze the overall performance over all 23 benchmark datasets using critical difference diagrams, illustrating the average ranks of the methods over all datasets and the rank statistics based on Wilcoxon-Holm with a significance level of $0.05$, sorting the methods based on their performance from right (best) to left (worst). The critical difference diagram follows the methodology of \citet{demsar2006statistical}, and the analysis is done using the average performance measured by the FSDEM~\citep{Rajabinasab2024fsdem} score for all of the metrics.

Figures~\ref{fg:line-rf} and \ref{fg:z-rf} present the results for the random forest classifier for all four evaluation metrics in the 10\% scenario. Clearly, the evaluation metrics assign a very good performance figure to the random baseline, despite its na\"ive approach and low computational complexity. On the Isolet dataset, SCFS and LLSRFS are the only methods performing better than the random baseline, while Variance Score and Laplacian Score are most of the time worse, VCSDFS and MCFS are worse, and SOGFS and the Correlation Score are the worst by a large margin.

In the critical difference diagrams (Figures~\ref{fg:cdd100-rf} and \ref{fg:cdd10-rf}), SCFS and MCFS are ranked first and second in the 100\% scenario, and swopped ranks in the 10\% scenario. The random baseline is ranked third in three of the ranking statistics and fifth for ACC over the full range, where Laplacian Score and Variance Score are ranked slightly ahead of it. However, none of the differences between SCFS, the random baseline, Variance Score, Laplacian Score, and VCSDFS are significant, and on a per-dataset level, SCFS outperforms the random baseline on only about half of the datasets. Let us note again that, moreover, SCFS and MCFS are not completely unsupervised, as they are using the number of classes as an input. The recent state-of-the-art method VCSDFS \citep{DBLP:journals/nn/KaramiSTMV23}, although also using information on the number of classes, is ranked behind the random baseline in all diagrams, and the Correlation is significantly worse than random over the first 10\% in terms of accuracy (ACC), as is LLSRFS in terms of AUC. SOGFS and LLSRFS are ranked behind the random baseline in all four diagrams as well. 

\perfresultsCD{RANDOMFOREST}{rf}{the random forest classifier}{ACC}{AUC}

It is noteworthy that the applied classifier, random forest, itself is built on random subsets of features and is robust to redundant features, reducing the gap between a carefully selected subset and a random one even more.
However, the results for KNN, decision tree, SVM, and XGBoost lead to the same conclusions (see Appendix~\ref{app:ext}). 
For all five classifiers, none of the methods is significantly better than the random baseline, and the Correlation is significantly worse than random in several of the diagrams, VCSDFS in two, and LLSRFS in one of them. The choice of the classifier mainly changes the relative order of the methods close to the random baseline. 
In conclusion, for supervised downstream tasks, none of the methods provides a feature subset which is considerably better than a randomly selected one.

\subsection{Unsupervised Performance}
The metrics CLSACC and NMI, based on an unsupervised downstream task (clustering), are evaluated using $k$-means clustering and do not depend on the classifier. The results on the Isolet dataset are shown in Figures~\ref{fg:line-unsup} and \ref{fg:z-unsup}. Similar to the supervised case, SCFS and LLSRFS are the only methods that are consistently better than the random baseline. Laplacian Score stays around the zero-line for CLSACC and is slightly above it for NMI, Variance Score is slightly below the random baseline, VCSDFS and MCFS are clearly below it, and SOGFS and the Correlation Score are far below it for both metrics. Therefore, the behavior of the methods on this dataset is largely independent of whether the downstream task is supervised or unsupervised.

The critical difference diagrams in Figures~\ref{fg:cdd100-unsup} and \ref{fg:cdd10-unsup} show a different picture compared to the supervised metrics. LLSRFS is ranked first in three of the four diagrams, and Laplacian Score in the remaining one and second in the others. The random baseline ranks third in three cases and fourth in the remaining case. However, none of the differences to the random baseline is significant. 
 
\perfresultsline{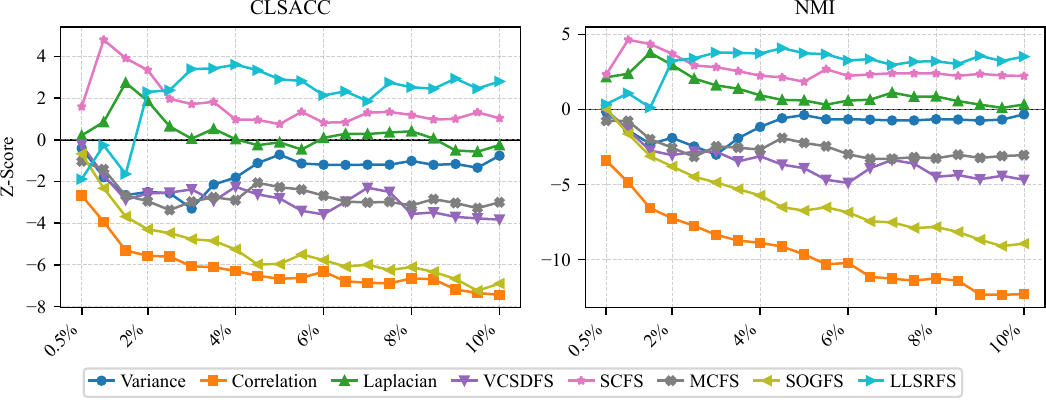}{unsup}{$k$-means clustering}{CLSACC}{NMI}
 
Laplacian Score selects the features which best preserve the local neighborhood structure of the data, which is closely aligned with the clustering objective. This explains its good ranks for the unsupervised metrics, and also why this advantage does not carry over to the supervised metrics. LLSRFS likewise learns the local structure of the data, and is ranked behind the random baseline in 19 of the 20 diagrams for the supervised metrics. Nevertheless, even in this favorable setting, the improvement over the random baseline is not significant, while the computational cost of Laplacian Score grows quadratically with the number of instances, and that of LLSRFS is among the highest of all methods. For the remaining methods, the unsupervised metrics confirm the supervised results, as none of them is able to outperform the random baseline consistently.

\subsection{Variability of the Random Baseline}
\begin{figure*}[tb!]
    \centering
    \includegraphics[width=\textwidth]{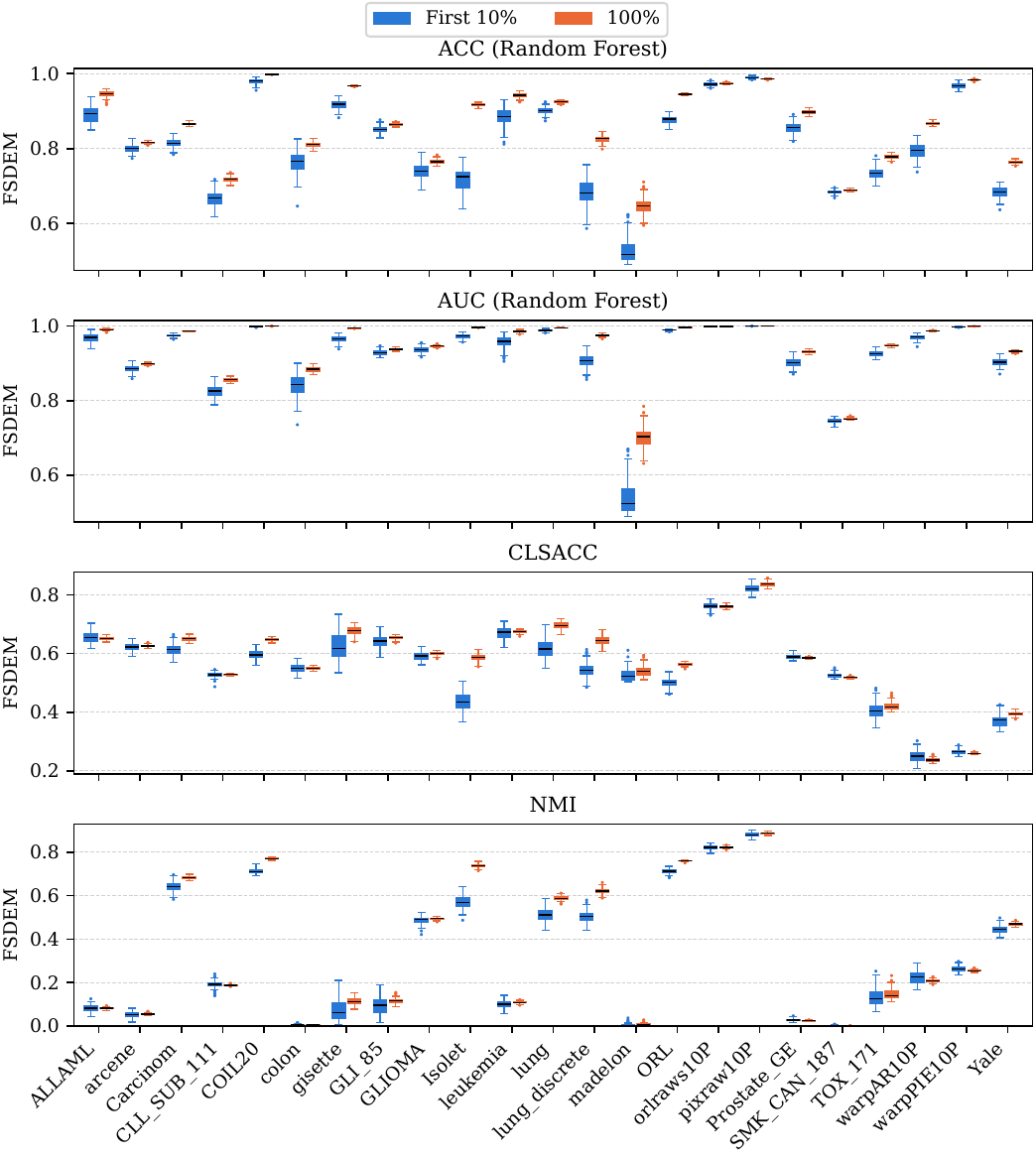}
    \Description{Four panels of box plots, one for each of ACC, AUC, CLSACC, and NMI, showing for each dataset the distribution of the FSDEM score of the random baseline over 100 runs, separately for the first 10\% and the 100\% experiment.}
    \caption{Distribution of the FSDEM score of the random baseline over 100 runs for each dataset, for ACC and AUC (using the random forest classifier), CLSACC, and NMI, for the first 10\% and the 100\% experiment. Boxes show the interquartile range and the median, whiskers extend to 1.5 times the interquartile range, and points mark outliers.}
    \label{fg:random-var}
\end{figure*}

The random baseline is not a single value, but a distribution over many independent selections. Fig.~\ref{fg:random-var} shows, for each dataset, the distribution of the performance of the random baseline over its 100 runs, for ACC and AUC (using the random forest classifier) as well as for CLSACC and NMI, and for both experiments (10\% and 100\%).

In most of the datasets, the spread of the random baseline is very small compared to its mean value, in particular over the full range. Hence, the good performance of the random baseline is not the result of a lucky draw, but a property of the datasets, in which almost any subset of features preserves enough information for the downstream task. In all datasets and for all metrics, the spread is larger over the first 10\% of features than over the full range, as the choice of the individual features becomes more important when only a few of them are selected. In this case, the difference between the best and the worst random run reaches up to about $0.2$ in FSDEM, which shows that a single random selection is not a reliable baseline, and that the mean and the standard deviation over many runs are required.

For the supervised metrics, the spread is closely linked to the performance level. Datasets on which the random baseline is close to the maximum performance, such as COIL20, pixraw10P, orlraws10P, and warpPIE10P, show almost no variation, as any subset of features is sufficient for the classifier. The lowest performance and one of the largest spreads are observed for madelon, a synthetic dataset designed with a small number of relevant features among many noise features. This is exactly the setting in which feature selection is expected to be beneficial, and indeed, Variance Score and Laplacian Score clearly exceed the spread of the random baseline on this dataset. For the unsupervised metrics, the spread is not related to the performance level, which reflects the sensitivity of $k$-means to the particular subset of features. Moreover, in contrast to the supervised metrics, selecting only a small fraction of features slightly improves the clustering performance on several datasets. On some datasets, such as colon, madelon, Prostate\_GE, and SMK\_CAN\_187, the NMI of the random baseline is close to zero, indicating that the clustering does not recover the class structure with any random subset. As most of the competing methods remain close to zero as well, NMI leaves very little room to distinguish between the feature selection methods on these datasets.

Relating the competing methods to this spread further supports our findings. Over all datasets, metrics, and both experiments, the FSDEM score of a competing method lies more than two standard deviations below the mean of the random baseline in 34\% of the cases and more than two standard deviations above it in only 21\% of the cases, and only LLSRFS exceeds this margin on more than half of the datasets, for CLSACC and NMI over the full range. Consequently, a competing method is more likely to be clearly worse than a random selection than to be clearly better than it.

\begin{figure*}[tbp!]
    \centering
    \includegraphics[width=0.92\textwidth]{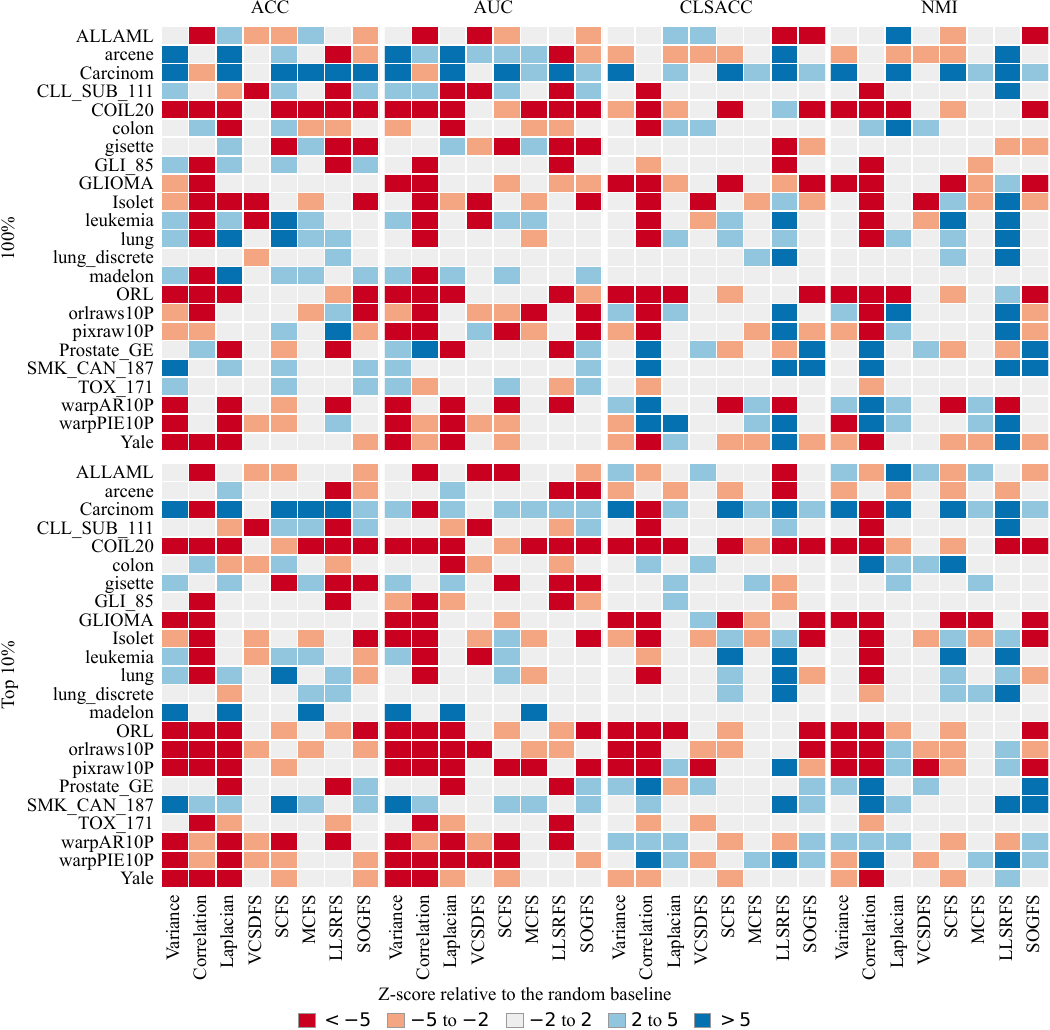}
    \Description{Eight heatmaps arranged in two rows and four columns, showing for each dataset and feature selection method the Z-score relative to the random baseline in five color-coded ranges, for ACC, AUC, CLSACC, and NMI, over the 100\% experiment and the first 10\% of features.}
    \caption{Z-score of the FSDEM score of each method relative to the random baseline on each dataset, for ACC, AUC, CLSACC, and NMI (left to right), over the 100\% experiment (top) and the first 10\% of features (bottom).}
    \label{fg:zscore-heatmap}
\end{figure*}

Figure~\ref{fg:zscore-heatmap} breaks this comparison down to the individual datasets. The patterns are very similar for the 100\% experiment and the first 10\% of features, as well as for ACC and AUC and for CLSACC and NMI. About two thirds of the cells which lie more than two standard deviations above the random baseline belong to gene expression datasets, while the image datasets COIL20, ORL, orlraws10P, pixraw10P, warpAR10P, warpPIE10P, and Yale reveal a clear weakness of several established methods. For the supervised metrics, Variance Score is below the random baseline on all seven image datasets in both experiments, and Correlation Score and Laplacian Score on most of them, mostly by more than five standard deviations, followed by SOGFS and SCFS, while VCSDFS and MCFS largely stay within the spread of the random baseline. A likely explanation is that neighboring pixels are highly correlated, so that these methods select many redundant pixels from the same image regions, whereas a random selection spreads over the whole image. For the unsupervised metrics, Laplacian Score and LLSRFS are not affected in the same way and lie within or above the spread of the random baseline on at least five of the seven image datasets. Moreover, the datasets on which one method clearly exceeds the random baseline are often those on which another method is clearly below it, which shows that none of the methods is uniformly better than a random selection.

\perfresultsCD{Unsupervised}{unsup}{$k$-means clustering}{CLSACC}{NMI}

\subsection{Discussion}
Clearly, the random baseline is among the best-performing unsupervised feature selection methods. In all presented critical difference diagrams, it is never significantly worse than any other method, and the same holds for the remaining classifiers.
Whenever a method is ranked ahead of the random baseline, the difference is less than two rank units and statistically insignificant. Even at a lenient significance level of $0.2$, only MCFS is significantly better than the random baseline, in a single setting (AUC over the first 10\% of features, using SVM). Moreover, the spread of the random baseline over its runs is small, and the competing methods are more often clearly worse than clearly better than random. Which methods are outperforming the random baseline depends on the downstream task, as SCFS and MCFS are the best methods for the supervised metrics and LLSRFS and Laplacian Score are the best methods for the unsupervised metrics, but none of them is consistently and significantly better than the random baseline in both settings. Many methods, such as the recent state-of-the-art methods VCSDFS \citep{DBLP:journals/nn/KaramiSTMV23} and SOGFS \citep{DBLP:conf/aaai/WuC21}, are consistently outperformed by the random baseline, putting in doubt whether their performance is actually acceptable at all. In combination with the efficiency analysis, where the random baseline is the most efficient method by a large margin, the additional computational cost of the competing methods is not justified by their performance. Overall, no method consistently and significantly outperforms the random baseline on both supervised and unsupervised downstream tasks, which clearly reflects the importance of a random baseline in the evaluation of unsupervised feature selection algorithms. 

The choice of datasets follows the common evaluation practice in the field. All 23 datasets belong to the scikit-feature repository \citep{fs_datasets}. This is the standard benchmark collection for feature selection and is also used in the evaluation of several of the competing methods \citep{DBLP:journals/eaai/ParsaZG20, DBLP:journals/nn/KaramiSTMV23, mcfs}. They are high-dimensional, and 19 of them contain more features than instances, which is precisely the setting in which feature selection is expected to be beneficial. Hence, the performance of the random baseline under the conditions typically applying to the development and comparison of unsupervised feature selection methods is a highly relevant finding.

\section{Conclusion}\label{sec:conc}
We discussed the importance of using a reasonable baseline for the evaluation of unsupervised feature selection methods. We reviewed the classic and recent state-of-the-art methods for unsupervised feature selection. We proposed using random feature selection as a baseline for the evaluation of unsupervised feature selection algorithms. We conducted extensive experiments on both classic and state-of-the-art feature selection methods to see whether they offer a significant improvement over random feature selection or not.

Let us note that the established evaluation measures reflect the performance of the classification or clustering task, and hence only indirectly the performance of the feature selection process as such. For example, using AUC, if the classifier randomly guesses the classes in a binary classification scenario, AUC equal to $0.5$ would be expected. However, a classifier is expected to perform better than random, even with a randomly selected subset of features. For ACC, CLSACC, and NMI, determining random behavior of the classifier or cluster assignment is not always equally straightforward, but in any case, what is measured is the \emph{downstream performance of the classifier or clustering method} on the selected features, \emph{not the selection of the features as such}. This might explain why close to (or worse than) random behavior of feature selection methods has not been noticed in the literature so far. \emph{Therefore, any supervised or unsupervised feature selection method should provide a feature subset considerably better than a randomly selected subset of the same size to be realistically deemed effective.}

The findings of this paper indicate that none of the evaluated methods significantly improves the performance of the unsupervised feature selection task in comparison with the random baseline. The random baseline is ranked among the top methods in most cases, mostly in second or third place. Whenever a method is performing better than the random baseline in some scenario and metric, there is no significant statistical difference observed, and the overall rank value is very close to the random baseline. Moreover, the performance of the random baseline varies little across independent runs, and the competing methods are more often clearly worse than random than clearly better than random, considering this variation. Also in terms of efficiency, as expected, the random baseline is superior, followed by the classic algorithms like variance-based and correlation-based methods. The evaluated recent methods showed a significantly higher computational complexity while offering no significant improvement over the random baseline.

This paper aims to emphasize the critical lack of a baseline for the evaluation of unsupervised feature selection algorithms. A baseline is required to guide the design and development of new methods, ensuring that the computational cost imposed by these algorithms is justified and that the proposed method can offer significantly better performance compared to the baseline and other approaches. We demonstrated that random feature selection can play the role of the baseline as it offers a consistently good downstream performance with almost no computational cost. In practice, we recommend reporting the random baseline as the mean and the standard deviation over many independent runs, and relating the competing methods to it by the Z-score and by statistical tests over multiple datasets, in particular when only a small number of features is selected. At the same time, our findings highlight a shortcoming in the common evaluation procedures for feature selection, unsupervised and supervised alike. As the traditional measures rely on downstream tasks and not on intrinsic properties of the features, they are not able to realize random-level quality of a feature selection method. 

For future work, developing new, advanced evaluation measures for feature selection remains a desideratum. 
In addition, our evaluation is based on a collection of datasets traditionally used as benchmark for feature selection and the findings showed that these datasets are not all useful as benchmark. It might therefore also be beneficial for future research in feature selection to develop a larger and more meaningful benchmark set.

\section*{Acknowledgments}
This study was funded by the Innovation Fund Denmark project ``PREPARE: Personalized Risk Estimation and Prevention of Cardiovascular Disease''.

\bibliographystyle{ACM-Reference-Format}
\bibliography{myref_dblp_condensed}
\appendix
\newcommand{\perfresultsds}[7]{%
\begin{figure*}[tp]
    \centering
    \includegraphics[width=0.8\textwidth]{figures/#1/line/10Percent/#6.pdf}
    \Description{Two line plots of #4 and #5 against the percentage of selected features on the #7 dataset, with one line per feature selection method and a shaded band for the standard deviation of the random baseline.}
    \caption{#4 and #5 of the unsupervised feature selection methods and the random baseline on the #7 dataset over the first 10\% of features, using #3. The shaded area shows the standard deviation of the random baseline.}
    \label{fg:line-#2-#6}
\end{figure*}
\begin{figure*}[tp]
    \centering
    \includegraphics[width=0.8\textwidth]{figures/#1/zscore/10Percent/#6.pdf}
    \Description{Two line plots of the Z-score of #4 and #5 relative to the random baseline against the percentage of selected features on the #7 dataset, with one line per feature selection method.}
    \caption{Z-score of #4 and #5 relative to the random baseline on the #7 dataset over the extreme dimensionality reduction experiment (0.5\% to 10\%), using #3.}
    \label{fg:z-#2-#6}
\end{figure*}}
\newcommand{\perfresultsdsfull}[7]{%
\begin{figure*}[tp]
    \centering
    \includegraphics[width=0.8\textwidth]{figures/#1/line/100Percent/#6.pdf}
    \Description{Two line plots of #4 and #5 against the percentage of selected features on the #7 dataset, with one line per feature selection method and a shaded band for the standard deviation of the random baseline.}
    \caption{#4 and #5 of the unsupervised feature selection methods and the random baseline on the #7 dataset over the full range of selected features (5\% to 100\%), using #3. The shaded area shows the standard deviation of the random baseline.}
    \label{fg:line-#2-#6-100}
\end{figure*}
\begin{figure*}[tp]
    \centering
    \includegraphics[width=0.8\textwidth]{figures/#1/zscore/100Percent/#6.pdf}
    \Description{Two line plots of the Z-score of #4 and #5 relative to the random baseline against the percentage of selected features on the #7 dataset, with one line per feature selection method.}
    \caption{Z-score of #4 and #5 relative to the random baseline on the #7 dataset over the full range of selected features (5\% to 100\%), using #3.}
    \label{fg:z-#2-#6-100}
\end{figure*}}

\newcommand{\apxcdd}[5]{%
\begin{figure*}[tp]
    \centering
    \includegraphics[width=0.49\textwidth]{cdd_output/#1/cdd_#4_100Percent_std.pdf}\hfill
    \includegraphics[width=0.49\textwidth]{cdd_output/#1/cdd_#5_100Percent_std.pdf}
    \Description{Two critical difference diagrams showing the average ranks of the feature selection methods for #4 and #5 over the 100\% experiment, with bars connecting methods whose differences are not significant.}
    \caption{Critical difference diagrams of #4 (left) and #5 (right) over the 100\% experiment based on the FSDEM score, using #3.}
    \label{fg:apx-cdd100-#2}
\end{figure*}
\begin{figure*}[tp]
    \centering
    \includegraphics[width=0.49\textwidth]{cdd_output/#1/cdd_#4_10Percent_std.pdf}\hfill
    \includegraphics[width=0.49\textwidth]{cdd_output/#1/cdd_#5_10Percent_std.pdf}
    \Description{Two critical difference diagrams showing the average ranks of the feature selection methods for #4 and #5 over the first 10\% of features, with bars connecting methods whose differences are not significant.}
    \caption{Critical difference diagrams of #4 (left) and #5 (right) over the first 10\% of features based on the FSDEM score, using #3.}
    \label{fg:apx-cdd10-#2}
\end{figure*}}

\newcommand{\apxdataset}[2]{%
\perfresultsds{KNN}{knn}{the KNN classifier}{ACC}{AUC}{#1}{#2}
\perfresultsds{DECISIONTREE}{dt}{the decision tree classifier}{ACC}{AUC}{#1}{#2}
\perfresultsds{SVM}{svm}{the SVM classifier}{ACC}{AUC}{#1}{#2}
\perfresultsds{RANDOMFOREST}{rf}{the random forest classifier}{ACC}{AUC}{#1}{#2}
\perfresultsds{XGBOOST}{xgb}{the XGBoost classifier}{ACC}{AUC}{#1}{#2}
\perfresultsds{Unsupervised}{unsup}{$k$-means clustering}{CLSACC}{NMI}{#1}{#2}
\clearpage}
\newcommand{\apxdatasetiso}[2]{%
\perfresultsds{KNN}{knn}{the KNN classifier}{ACC}{AUC}{#1}{#2}
\perfresultsds{DECISIONTREE}{dt}{the decision tree classifier}{ACC}{AUC}{#1}{#2}
\perfresultsds{SVM}{svm}{the SVM classifier}{ACC}{AUC}{#1}{#2}
\perfresultsds{XGBOOST}{xgb}{the XGBoost classifier}{ACC}{AUC}{#1}{#2}
\clearpage}
\newcommand{\apxdatasetfull}[2]{%
\perfresultsdsfull{KNN}{knn}{the KNN classifier}{ACC}{AUC}{#1}{#2}
\perfresultsdsfull{DECISIONTREE}{dt}{the decision tree classifier}{ACC}{AUC}{#1}{#2}
\perfresultsdsfull{SVM}{svm}{the SVM classifier}{ACC}{AUC}{#1}{#2}
\perfresultsdsfull{RANDOMFOREST}{rf}{the random forest classifier}{ACC}{AUC}{#1}{#2}
\perfresultsdsfull{XGBOOST}{xgb}{the XGBoost classifier}{ACC}{AUC}{#1}{#2}
\perfresultsdsfull{Unsupervised}{unsup}{$k$-means clustering}{CLSACC}{NMI}{#1}{#2}
\clearpage}

\newcommand{\apxfig}[2]{\hyperref[fg:#1-#2]{Fig.~\ref*{fg:#1-#2}}}
\newcommand{\apxcells}[2]{\apxfig{line-#1}{#2} & \apxfig{z-#1}{#2}}
\newcommand{\apxrow}[2]{\hyperref[fg:line-knn-#1]{#2} & \apxcells{knn}{#1} & \apxcells{dt}{#1} & \apxcells{svm}{#1} & \apxcells{rf}{#1} & \apxcells{xgb}{#1} & \apxcells{unsup}{#1} \\}
\newcommand{\apxfigmain}[1]{\hyperref[fg:#1]{Fig.~\ref*{fg:#1}}}
\newcommand{\apxrowiso}[2]{\hyperref[fg:line-knn-#1]{#2} & \apxcells{knn}{#1} & \apxcells{dt}{#1} & \apxcells{svm}{#1} & \apxfigmain{line-rf} & \apxfigmain{z-rf} & \apxcells{xgb}{#1} & \apxfigmain{line-unsup} & \apxfigmain{z-unsup} \\}
\newcommand{\apxrowfull}[2]{\hyperref[fg:line-knn-#1-100]{#2} & \apxcells{knn}{#1-100} & \apxcells{dt}{#1-100} & \apxcells{svm}{#1-100} & \apxcells{rf}{#1-100} & \apxcells{xgb}{#1-100} & \apxcells{unsup}{#1-100} \\}

\clearpage
\twocolumn[{%
\section{Extended Experimental Results}\label{app:ext}
In the main text, we show the evaluation results on the Isolet dataset over the first 10\% of features as a representative example, for the random forest classifier and for $k$-means clustering. Here, we show the extended experimental results on all datasets included in the study, for both the first 10\% of features (0.5\% to 10\%) and the full range of selected features (5\% to 100\%). For every dataset, we present the performance of the feature selection methods and the random baseline, and the corresponding Z-scores relative to the random baseline. The supervised metrics (ACC and AUC) are shown for all five classifiers, namely KNN, decision tree, SVM, random forest, and XGBoost, and the unsupervised metrics (CLSACC and NMI) are shown for $k$-means clustering. The critical difference diagrams aggregate the results over all datasets. They are shown first for the KNN, decision tree, SVM, and XGBoost classifiers (Figures~\ref{fg:apx-cdd100-knn} to~\ref{fg:apx-cdd10-xgb}), as the diagrams for the random forest classifier and for $k$-means clustering are provided in the main text. Tables~\ref{tbl:apx-nav10} and~\ref{tbl:apx-nav100} list all figures, and every entry links to the corresponding figure. These are followed by the performance figures and Z-score figures for the first 10\% of features and then for the full range. For Isolet over the first 10\% of features, the figures for the random forest classifier and for $k$-means clustering are shown in the main text and are therefore not repeated here. The corresponding entries in Table~\ref{tbl:apx-nav10} link to the main text.

\vspace{4pt}
\begin{center}
    \refstepcounter{table}\label{tbl:apx-nav10}
    {\small\textbf{Table~\thetable: Overview of the extended results for the first 10\% of features (the results on Isolet for the random forest classifier and for $k$-means are shown in the main text): ACC and AUC for the five classifiers and CLSACC and NMI for $k$-means. Each entry links to the corresponding figure.}\par}
    \vspace{4pt}
    \footnotesize
    \setlength{\tabcolsep}{3.5pt}
    \begin{tabular}{lcccccccccccc}
    \toprule
    & \multicolumn{2}{c}{KNN} & \multicolumn{2}{c}{Decision tree} & \multicolumn{2}{c}{SVM} & \multicolumn{2}{c}{Random forest} & \multicolumn{2}{c}{XGBoost} & \multicolumn{2}{c}{$k$-means} \\
    \cmidrule(lr){2-3} \cmidrule(lr){4-5} \cmidrule(lr){6-7} \cmidrule(lr){8-9} \cmidrule(lr){10-11} \cmidrule(lr){12-13}
    Dataset & Line & Z-score & Line & Z-score & Line & Z-score & Line & Z-score & Line & Z-score & Line & Z-score \\
    \midrule
        \apxrow{ALLAML}{ALLAML}
        \apxrow{arcene}{arcene}
        \apxrow{Carcinom}{Carcinom}
        \apxrow{CLL_SUB_111}{CLL\_SUB\_111}
        \apxrow{COIL20}{COIL20}
        \apxrow{colon}{colon}
        \apxrow{gisette}{gisette}
        \apxrow{GLI_85}{GLI\_85}
        \apxrow{GLIOMA}{GLIOMA}
        \apxrowiso{Isolet}{Isolet}
        \apxrow{leukemia}{leukemia}
        \apxrow{lung}{lung}
        \apxrow{lung_discrete}{lung\_discrete}
        \apxrow{madelon}{madelon}
        \apxrow{ORL}{ORL}
        \apxrow{orlraws10P}{orlraws10P}
        \apxrow{pixraw10P}{pixraw10P}
        \apxrow{Prostate_GE}{Prostate\_GE}
        \apxrow{SMK_CAN_187}{SMK\_CAN\_187}
        \apxrow{TOX_171}{TOX\_171}
        \apxrow{warpAR10P}{warpAR10P}
        \apxrow{warpPIE10P}{warpPIE10P}
        \apxrow{Yale}{Yale}
    \bottomrule
    \end{tabular}
\end{center}
\vspace{6pt}
}]

\twocolumn[{%
\vspace{4pt}
\begin{center}
    \refstepcounter{table}\label{tbl:apx-nav100}
    {\small\textbf{Table~\thetable: Overview of the extended results for the full range of selected features: ACC and AUC for the five classifiers and CLSACC and NMI for unsupervised ($k$-means). Each entry links to the corresponding figure.}\par}
    \vspace{4pt}
    \footnotesize
    \setlength{\tabcolsep}{3.5pt}
    \begin{tabular}{lcccccccccccc}
    \toprule
    & \multicolumn{2}{c}{KNN} & \multicolumn{2}{c}{Decision tree} & \multicolumn{2}{c}{SVM} & \multicolumn{2}{c}{Random forest} & \multicolumn{2}{c}{XGBoost} & \multicolumn{2}{c}{$k$-means} \\
    \cmidrule(lr){2-3} \cmidrule(lr){4-5} \cmidrule(lr){6-7} \cmidrule(lr){8-9} \cmidrule(lr){10-11} \cmidrule(lr){12-13}
    Dataset & Line & Z-score & Line & Z-score & Line & Z-score & Line & Z-score & Line & Z-score & Line & Z-score \\
    \midrule
        \apxrowfull{ALLAML}{ALLAML}
        \apxrowfull{arcene}{arcene}
        \apxrowfull{Carcinom}{Carcinom}
        \apxrowfull{CLL_SUB_111}{CLL\_SUB\_111}
        \apxrowfull{COIL20}{COIL20}
        \apxrowfull{colon}{colon}
        \apxrowfull{gisette}{gisette}
        \apxrowfull{GLI_85}{GLI\_85}
        \apxrowfull{GLIOMA}{GLIOMA}
        \apxrowfull{Isolet}{Isolet}
        \apxrowfull{leukemia}{leukemia}
        \apxrowfull{lung}{lung}
        \apxrowfull{lung_discrete}{lung\_discrete}
        \apxrowfull{madelon}{madelon}
        \apxrowfull{ORL}{ORL}
        \apxrowfull{orlraws10P}{orlraws10P}
        \apxrowfull{pixraw10P}{pixraw10P}
        \apxrowfull{Prostate_GE}{Prostate\_GE}
        \apxrowfull{SMK_CAN_187}{SMK\_CAN\_187}
        \apxrowfull{TOX_171}{TOX\_171}
        \apxrowfull{warpAR10P}{warpAR10P}
        \apxrowfull{warpPIE10P}{warpPIE10P}
        \apxrowfull{Yale}{Yale}
    \bottomrule
    \end{tabular}
\end{center}
\vspace{6pt}
}]

\apxcdd{KNN}{knn}{the KNN classifier}{ACC}{AUC}
\apxcdd{DECISIONTREE}{dt}{the decision tree classifier}{ACC}{AUC}
\apxcdd{SVM}{svm}{the SVM classifier}{ACC}{AUC}
\apxcdd{XGBOOST}{xgb}{the XGBoost classifier}{ACC}{AUC}
\clearpage

\apxdataset{ALLAML}{ALLAML}
\apxdataset{arcene}{arcene}
\apxdataset{Carcinom}{Carcinom}
\apxdataset{CLL_SUB_111}{CLL\_SUB\_111}
\apxdataset{COIL20}{COIL20}
\apxdataset{colon}{colon}
\apxdataset{gisette}{gisette}
\apxdataset{GLI_85}{GLI\_85}
\apxdataset{GLIOMA}{GLIOMA}
\apxdatasetiso{Isolet}{Isolet}
\apxdataset{leukemia}{leukemia}
\apxdataset{lung}{lung}
\apxdataset{lung_discrete}{lung\_discrete}
\apxdataset{madelon}{madelon}
\apxdataset{ORL}{ORL}
\apxdataset{orlraws10P}{orlraws10P}
\apxdataset{pixraw10P}{pixraw10P}
\apxdataset{Prostate_GE}{Prostate\_GE}
\apxdataset{SMK_CAN_187}{SMK\_CAN\_187}
\apxdataset{TOX_171}{TOX\_171}
\apxdataset{warpAR10P}{warpAR10P}
\apxdataset{warpPIE10P}{warpPIE10P}
\apxdataset{Yale}{Yale}

\apxdatasetfull{ALLAML}{ALLAML}
\apxdatasetfull{arcene}{arcene}
\apxdatasetfull{Carcinom}{Carcinom}
\apxdatasetfull{CLL_SUB_111}{CLL\_SUB\_111}
\apxdatasetfull{COIL20}{COIL20}
\apxdatasetfull{colon}{colon}
\apxdatasetfull{gisette}{gisette}
\apxdatasetfull{GLI_85}{GLI\_85}
\apxdatasetfull{GLIOMA}{GLIOMA}
\apxdatasetfull{Isolet}{Isolet}
\apxdatasetfull{leukemia}{leukemia}
\apxdatasetfull{lung}{lung}
\apxdatasetfull{lung_discrete}{lung\_discrete}
\apxdatasetfull{madelon}{madelon}
\apxdatasetfull{ORL}{ORL}
\apxdatasetfull{orlraws10P}{orlraws10P}
\apxdatasetfull{pixraw10P}{pixraw10P}
\apxdatasetfull{Prostate_GE}{Prostate\_GE}
\apxdatasetfull{SMK_CAN_187}{SMK\_CAN\_187}
\apxdatasetfull{TOX_171}{TOX\_171}
\apxdatasetfull{warpAR10P}{warpAR10P}
\apxdatasetfull{warpPIE10P}{warpPIE10P}
\apxdatasetfull{Yale}{Yale}
\end{document}